\documentclass[conference]{IEEEtran}
\IEEEoverridecommandlockouts
\usepackage{array}
\usepackage{float}
\usepackage{multirow}
\usepackage{amsmath}
\usepackage{amssymb}
\usepackage{graphicx}
\usepackage{balance}
\usepackage{cite}
\usepackage{amsmath,amssymb,amsfonts}
\usepackage{graphicx}
\usepackage{textcomp}
\usepackage{xcolor}
\usepackage[linesnumbered,ruled]{algorithm2e}
\usepackage{algorithmicx}

\makeatletter

\begin{document}
\title{Path Planning for Shepherding a Swarm in a Cluttered Environment using Differential Evolution}

\author{\IEEEauthorblockN{Saber Elsayed\IEEEauthorrefmark{1}, Hemant Singh\IEEEauthorrefmark{1}, Essam Debie\IEEEauthorrefmark{1}, Anthony Perry\IEEEauthorrefmark{2}, Benjamin Campbell\IEEEauthorrefmark{2}, Robert Hunjet\IEEEauthorrefmark{2}, Hussein Abbass\IEEEauthorrefmark{1}}
\IEEEauthorblockA{\IEEEauthorrefmark{1}School of Engineering and Information Technology, University of New South Wales, Canberra ACT, Australia. \\ Emails:
\{s.elsayed,h.singh,e.debie,h.abbass\}@unsw.edu.au}
\IEEEauthorblockA{\IEEEauthorrefmark{2}Defence Science and Technology, Australian Department of Defence, Edinburgh SA, Australia. \\ Emails:
\{Anthony.Perry,Benjamin.Campbell,Robert.Hunjet\}@dst.defence.gov.au } } 
\maketitle
\begin{abstract}
Shepherding involves herding a swarm of agents~(\emph{sheep}) by another a control agent~(\emph{sheepdog}) towards a goal. 
Multiple approaches have been documented in the literature to model this behaviour. In this paper, we present a modification to a well-known shepherding approach, and show, via simulation, that this modification improves shepherding efficacy. We then argue that given complexity arising from obstacles laden environments, path planning approaches could further enhance this model. To validate this hypothesis, we present a 2-stage evolutionary-based path planning algorithm for shepherding a swarm of agents in 2D environments. In the first stage, the algorithm attempts to find the best path for the sheepdog to move from its initial location to a strategic driving location behind the sheep. In the second stage, it calculates and optimises a path for the sheep. It does so by using \emph{way points} on that path as the sequential sub-goals for the sheepdog to aim towards. The proposed algorithm is evaluated in obstacle laden environments via simulation with further improvements achieved. 
\end{abstract}


\section{Introduction}
\label{sec:introduction}

Shepherding refers to the management and guidance of a sheep herd using one or more sheepdogs. In computational intelligence research, the concept is used more broadly to model and analyze the behaviour of biologically inspired swarms, where multiple agents of different type interact with each other in a proactive and reactive manner. The reactive agents are analogous to the sheep in the problem; they respond to the presence of the proactive agent, the sheepdog, and are repulsed from it. The sheepdog makes a sequence of decisions to influence the sheep and to guide them towards a goal area. A recent comprehensive review on the subject can be found in~\cite{long2020comprehensive}. The shepherding problem using robotic swarms is of interest in several applications beyond the biological inspiration of shepherding itself; applications include  crowd control~\cite{lien2009interactive}, clean-up of oil spills~\cite{fingas2012basics}, disaster relief and rescue operations~\cite{shell2004directional}, and security/military procedures~\cite{chaimowicz2007aerial}, among others. 

The shepherding problem shares some similarities with problem of efficient navigation of mobile robots, where path planning has been widely studied in the literature. A recent review of such methods for path planning appears in~\cite{patle2019review}. The navigation may be global or local in nature. The former assumes complete information of the environment a priori and creates an efficient path to move towards a pre-defined goal. The latter is more concerned with dynamic changes based on the relative positions among various elements. They are also referred to as reactive approaches as they consider the environment to adapt their path and are able to navigate autonomously. Correspondingly, the approaches for path planning can be broadly categorized into two classes: classical and reactive~\cite{patle2019review}. 

Evidently, the path planning problem becomes more difficult to solve with an increasing number of agents navigating the environment, as well as an increasing number of obstacles in the environment, among other factors. With shepherding, as opposed to a generic navigation, the problem is further complicated since the path planning algorithm needs to consider two different types of agents~(sheepdog and sheep). The sheepdog, which acts as the controlling agent in the problem, needs to consider the optimisation of its own path towards a driving point, as well as clustering the sheep (noting that they behave with their own dynamics) into a flock and driving them towards the goal, while negotiating the other obstacles in the environment. 




In this paper, we present an evolutionary path planning approach for shepherding that takes into account the collection and movement of the swarm~(sheep) in addition to the sheepdog. The problem is different from conventional path planning for robot navigation in the sense that the control agents~(sheepdog) have access to global information when seeking an optimal path, while the movement of others~(sheep) is purely reactive. The two-phase algorithm starts by identifying the path for the sheepdog to move from any initial position to a position behind the swarm. The path is constrained to be obstacle free and so as not to impact the sheep; lest the sheep be repulsed and scatter, making their collection even harder and more time-consuming. In the second phase, the algorithm plans the path for the sheepdog by identifying the next series of way points to guide the sheep towards their final destination.

We identify from the related work~(given in Section~\ref{sec:background}) that the potential of evolutionary algorithms for solving path planning problems in shepherding has not been significantly explored. In particular, we are interested in shepherding in environments cluttered by several obstacles. For this study, we assume that there is only one sheepdog and multiple sheep. Along with collision avoidance, the complex interplay of forces between the sheepdog and sheep makes the task even more challenging, for which the classical techniques are not geared. 

We propose the use of a differential evolution~(DE) based approach to path planning within the shepherding problem. The key contributions of this study are as follows:
\begin{itemize}
    \item improvement in the baseline Str\"{o}mbom model itself by considering the  positions of the isolated sheep relative to the flock, in order to decide on collecting/driving behavior.  
    \item the introduction of a two stage path planner within the shepherding problem addressing how the shepherd approaches the driving point whilst not disturbing the flock, and how it can then drive the flock to the goal,
    \item the ability to plan for driving using an evolutionary approach whilst taking into account obstacles that the flock may encounter (note, from a scalability perspective the path needs only to be calculated by the shepherd, not by any of the sheep), and
    \item validating the above contributions and the efficacy via simulation, showing that our DE path planning approach can indeed benefit shepherding.
\end{itemize}

In the remainder of this paper, we first provide a brief outline of selected works from the literature in Section~\ref{sec:background}. The proposed algorithm is presented in detail in Section~\ref{sec:algorithm}, followed by experimental design and results in Section~\ref{sec:experiments}. Concluding remarks are given in Section~\ref{sec:conclusion}.

\section{Related work}
\label{sec:background}

In this section, we review some of the existing work relevant to this study, focusing on three key aspects - shepherding in general, path planning and the use of computational intelligence methods in the context of these problems. 

\subsection{Shepherding}

The inspiration for modelling of shepherding can be traced back to the study of animal behaviours in~\cite{reynolds1987flocks}. The development of robotic shepherd was then formalised in subsequent research such as~\cite{vaughan1997introducing,vaughan2000experiments}. Since then, a number of works have been conducted on the topic, including the simulation and analysis of the swarming behaviours, guidance strategies, and implementation on real systems, most of which are reviewed in~\cite{long2020comprehensive}.

A shepherding problem can be initialised with a defined boundary of the field, a defined goal area to which the sheep need to be driven to, the initial position(s) of sheepdog and sheep, and the obstacles~(if any) that are present in the field~\cite{el2020preliminary}. An obstacle refers to a region in the field that is inaccessible to the agents. Moreover, depending on the problem of interest, the behaviour of the agents, as well as their capabilities can be defined. For example, if shepherding is done through unmanned aerial vehicles~(UAVs), then they can negotiate the ground obstacles by simply flying over them instead of going around them.  

Starting from the initial state, the sheepdog\textquoteright s aim is to group the sheep in a cohesive flock and guide them towards a goal. The primitive behaviours are defined as  \emph{collecting} and \emph{driving} in a paper by Str{\"o}mbom et~ al\cite{strombom2014solving}. The relative movement of the sheepdog and sheep are modelled using attraction-repulsion forces, which are calculated using various weighting factors including those for collision avoidance, inertia, attraction towards the centre of mass, and natural Jittering movements. 

\subsection{Path planning}

Existing path planning approaches for mobile robot navigation are classified into two types: classical and reactive~\cite{patle2019review}. We discuss these below. 

\subsubsection{Classical path planning approaches} 
Some examples of classical path planning approaches include: 
\begin{itemize}
    \item Cell decomposition (CD) approach: It divides the search space into a number of non-overlapping cells, with the starting and final positions assigned to specific cells~\cite{vseda2007roadmap,glavavski2009robot}. The path is constructed using a sequence of connected cells that do not contain an obstacle. The cells containing an obstacle could then be further split into smaller sizes to create feasible candidate cells for construction of a more efficient path. The cell shapes are often considered to be regular~(e.g. grid with square cells), but a number of approaches also consider irregular shapes.  
    
    \item Roadmap approach (RA): The roadmap approach resembles a graph whereby the connections~(edges) between different points~(nodes) can be traversed to construct the path. Voronoi diagrams and visibility graphs are some of the commonly used techniques to construct the roadmap~\cite{bhattacharya2008roadmap,wein2007visibility}.
    
    \item Artificial potential field (APF) approach: In this approach, a potential field is created in the search region by considering the agents and obstacles akin to charged particles~\cite{khatib1986real,huang2008velocity}. The forces exerted on the robot cause attraction towards the goal and repulsion from the obstacles. A feasible path can thus be constructed via the resultant field.
\end{itemize}

\subsubsection{Reactive path planning approaches}     

One of the perceived limitations of classical approaches is that they are not suitable for real-time applications involving uncertainty. The use of reactive path planning approaches has been proposed in the literature to provide more robust solutions in such cases. The reactive approaches do not assume global knowledge and instead make decisions based on local sensory data to execute in effect a \textquoteleft mapless\textquoteright \  navigation. A range of reactive methods is surveyed in~\cite{hoy2015algorithms}, with a particular focus on model predictive and sliding mode control. The techniques for collision avoidance include potential field methods for moving obstacles, reciprocal methods, and hybrid logic approaches. In~\cite{patle2019review}, the focus of the reactive approaches reviewed was on the use of metaheuristics to solve the underlying optimisation and modelling problems of the mobile agents. A number of such techniques have been applied in this context, such as genetic algorithms~\cite{shing1993genetic,shi2010dynamic,kang2011genetic}, fuzzy logic~\cite{castellano1997automatic}, neural networks~\cite{na2003hybrid}, and swarm intelligence based search methods~\cite{tang2014mobile,liu2019comprehensive,zhang2013robot}.   

Even though path planning has been a well-researched area in general for multi-agent robotic systems, the works focusing specifically on shepherding have been relatively few. A roadmap based approach was studied in~\cite{bayazit2002roadmap} for shepherding in complex environments and shown to perform better than a bitmap~(cell decomposition based) approach. In~\cite{lien2004shepherding} the path of the sheepdog was modelled based on the current state of shepherding~(approach or steer). While the former involves moving in straight lines, safe zones, or dynamic roadmaps, the latter involves positioning directly behind, side-to-side movement or turning the flock. The roadmap approach was also further extended to deal with more diverse environments in~\cite{vo2009behavior}. In~\cite{harrison2010scalable}, the sheep were considered collectively as deformable shapes instead of individual agents to improve scalability and robustness of the shepherding task. And finally, in~\cite{longshepherding}, a circular path was enforced so that the sheepdog does not split the flock while moving to the driving position. 

\section{Shepherding model}\label{sec:algorithm}

In this section, we describe the details of the shepherding models we use for this study. 
Our approaches utilise the model proposed 
by Str\"{o}mbom et al.~\cite{strombom2014solving}, discussed in Section~\ref{subsec:strombom}. However, we propose an improvement in the algorithm, we call UNSWDST, discussed in Section~\ref{subsec:unswdst}. After demonstrating the improvements gained by this approach, we further utilise it as a baseline for demonstration of the efficiency gains which can be afforded by our DE path planning method.

\subsection{Str\"{o}mbom et al's algorithm}\label{subsec:strombom}

The Str\"{o}mbom et al.~\cite{strombom2014solving} model, governs the dynamics of the sheepdog and sheep, by defining the specific manners in which they interact with each other and the environment. In principle, the movement of sheep is represented as a weighted linear combination of \textit{force vectors} which represent the influence of various entities on the sheep.    

The set of sheep agents are denoted by $\Pi = \{ \pi_1, \dots, \pi_i, \dots, \pi_N \}$, while the shepherd (sheepdog) agents denoted as $B = \{ \beta_1, \dots, \beta_j, \dots, \beta_M \}$, the set of behaviours in the simulation with $\Sigma = \{ \sigma_1, \dots, \sigma_K \}$, and the set of obstacles as $O=\{O_1, O_2, ..., O_{N} \}$. In our model, the obstacles induce a repulsive force on the sheep in a similar manner in that the sheep are repulsed by the sheepdog. As per~\cite{strombom2014solving, singh2019modulation}, the agents adopt different behaviors:

\begin{enumerate}

\item Driving behavior: When the sheep is clustered in one group, the shepherd drives the sheep towards the goal by moving towards a driving point that is situated behind the sheep on the ray between the sheep center of mass and the goal. The shepherd moves towards the driving point with normalized force vector, $F^t_{\beta_jcd}$.

\item Collecting behavior: If one of the sheep is further away from the group, the sheepdog drives to a collection point (with a normalized force vector $F^t_{\beta_jcd}$) behind this sheep to move it to the herd.  
\item Sheepdog $\beta_j$ adds a random force, $F^t_{\beta_j\epsilon}$, at each time-step to help resolve deadlocks. The strength of this angular noise is denoted by $W_{e\beta_j}$.

\item Sheepdog $\beta_j$ total force $F^t_{\beta_j}$ is then calculated as:

\begin{equation}\label{eq:ShepherdTotalForceEquation} F^t_{\beta_j} = F^t_{\beta_jcd} + W_{e\beta_j}  F^t_{\beta_j\epsilon} \end{equation}

\item Sheep $\pi_i$ is repulsed from sheepdog $\beta_j$ using a force  $F^t_{\pi_i\beta}$. 

\item Sheep $\pi_i$ is repulsed from other sheep $\pi_{i1}, {i_1} \ne i$ using a force $F^t_{\pi_i\pi_{i1}}$.

\item Sheep $\pi_i$ is attracted to the center of mass of its neighbors $\Lambda^t_{\pi_i}$ using a force $F^t_{\pi_i\Lambda^t_{\pi_i}}$. 

\item Sheep $\pi_i$ angular noise uses a force $F^t_{\pi_i\epsilon}$. 

\item Sheep $\pi_i$ total force is calculated as:

\[ F^t_{\pi_i}  = W_{\pi_\upsilon}  F^{t-1}_{\pi_i} +
W_{\pi \Lambda} F^t_{\pi_i{\Lambda^t_{\pi_i}}} + W_{\pi\beta} F^t_{\pi_i\beta_j}  +\]
\begin{equation}\label{eq:SheepTotalForceEquation} W_{\pi\pi}  F^t_{\pi_i\pi_{-i}}  + W_{e\pi_i} F^t_{\pi_i\epsilon} 
\end{equation}

where each $W$ representing the weight of the corresponding force vector.

\end{enumerate}

The total force of each agent is used to update the agent position as depicted in Equations~\ref{eq:UpdatedSheepPositionEquation} and~\ref{eq:UpdatedShepherdPositionEquation}. If there is a sheep within three times the sheep-to-sheep interaction radius, the agent will stop; thus, it will set its speed to zero: $S^t_{\beta_j} = 0$, otherwise it will use its default speed, $S^t_{\beta_j} = S_{\beta_j}$. The speed of a sheep is assumed constant; that is, $S^t_{\pi_i} = S_{\pi_i}$

\begin{equation}\label{eq:UpdatedSheepPositionEquation}
P^{t+1}_{\pi_i} = P^t_{\pi_i} + S^t_{\pi_i} F^t_{\pi_i}
\end{equation}

\begin{equation}\label{eq:UpdatedShepherdPositionEquation}
P^{t+1}_{\beta_j} = P^{t}_{\beta_j} + S^t_{\beta_j} F^t_{\beta_j}
\end{equation}

Both $\pi$ and $\beta$ agents in Str\"{o}mbom model move with fixed speed. Generally, this is neither biologically plausible, since sheepdogs, for example, do not move with a constant speed during shepherding, nor technologically appropriate considering vehicle dynamics. Constant speed also limits the ability for the model to encapsulate behavioural attributes pertaining to closing speed  (e.g., aggressiveness) that can influence both the effectiveness and efficiency of shepherding ~\cite{singh2019modulation}.

\subsection{Modified algorithm~(UNSWDST)} 
\label{subsec:unswdst}
\begin{figure}
    
     \includegraphics[width=0.5\textwidth]{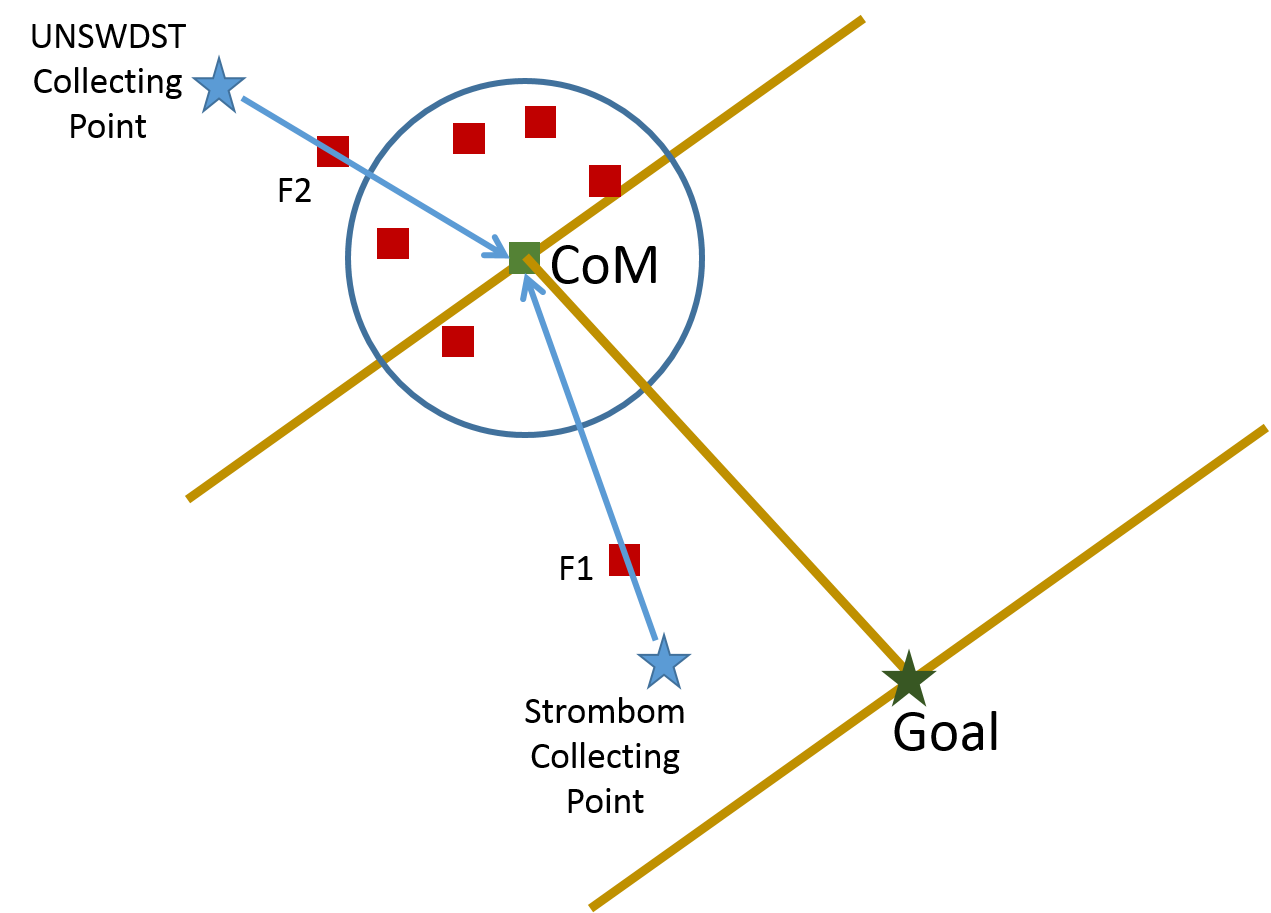}
    \caption{The difference between Str\"{o}mbom et al and UNSWDST shepherding algorithm.}
    \label{fig:unswdst}
\end{figure}

When analysing Str\"{o}mbom et al\textquoteright s original algorithm, it was apparent that the sheepdog lacked the intuition of its biological counterpart. In particular, Figure~\ref{fig:unswdst} shows two sheep outside the flock range: $F1$ and $F2$. According to Str\"{o}mbom et al, $F1$ is the furthest sheep, and should be collected. However, it could be inefficient to proceed to collect $F1$ since the herd will encounter $F1$ as it approaches the goal, and the natural attraction force between $F1$ and the members of the herd will provide cohesion. As such, we discount the furthest sheep if it is located between the two brown parallel lines in  Figure~\ref{fig:unswdst}. These two parallel lines are both perpendicular on the vector between the Centre of Mass (CoM) and the goal. Only sheep outside the herd zone and outside the area between these two lines are considered in the calculation of furthest sheep. In the situation depicted in  Figure~\ref{fig:unswdst}, Str\"{o}mbom et al\textquoteright would move to collect $F1$, while the UNSWDST algorithm will select $F2$.

\section{Methodology}
This section discusses the framework and key components of the proposed methodology for path planning.
\subsection{Two-Phase Framework For Sheep Driving and Collision Avoidance}
Our framework offers a two-phase solution combining collision-free path planning for both the shepherd to reach the driving position and then for the flock to be driven to the designated goal. Particularly, within the first stage, Differential Evolution~(DE) is used to optimise the shortest path to move the shepherd to a driving point which does not adversely affect the flock, while in the second stage, the shortest path from the flock's $GCM$ to target position is optimised to drive the sheep whilst avoiding collision with obstacles in the environment.  

\begin{algorithm}[phbt] 
\begin{algorithmic}[1]

\State $P_{\beta_j}$ $\leftarrow$ current position of the shepherd $\beta_j$ 
\State $P_{d_j}$ $\leftarrow$ the target driving position for shepherd $\beta_j$ 
\State $A^*_{\beta}$ $\leftarrow$ ordered list of waypoints calculated using spline interpolation along the optimised path from $P_{\beta_j}$ to $P_d$
\State  $A^*_{goal}$ $\leftarrow$ ordered list of sub-goals 

\While {$A^*_{goal}$ is not empty}
{
    \State $p_i$ $\leftarrow$ the next way point in $A^*_{goal}$ \\
    \State Drive the flock to $p_i$ with fixed speed 
}
  \caption{Two-stage shepherding framework} \label{alg:framework} 
   \end{algorithmic}
  \end{algorithm}

\begin{figure}
     \centering
     \includegraphics[width=0.45\textwidth]{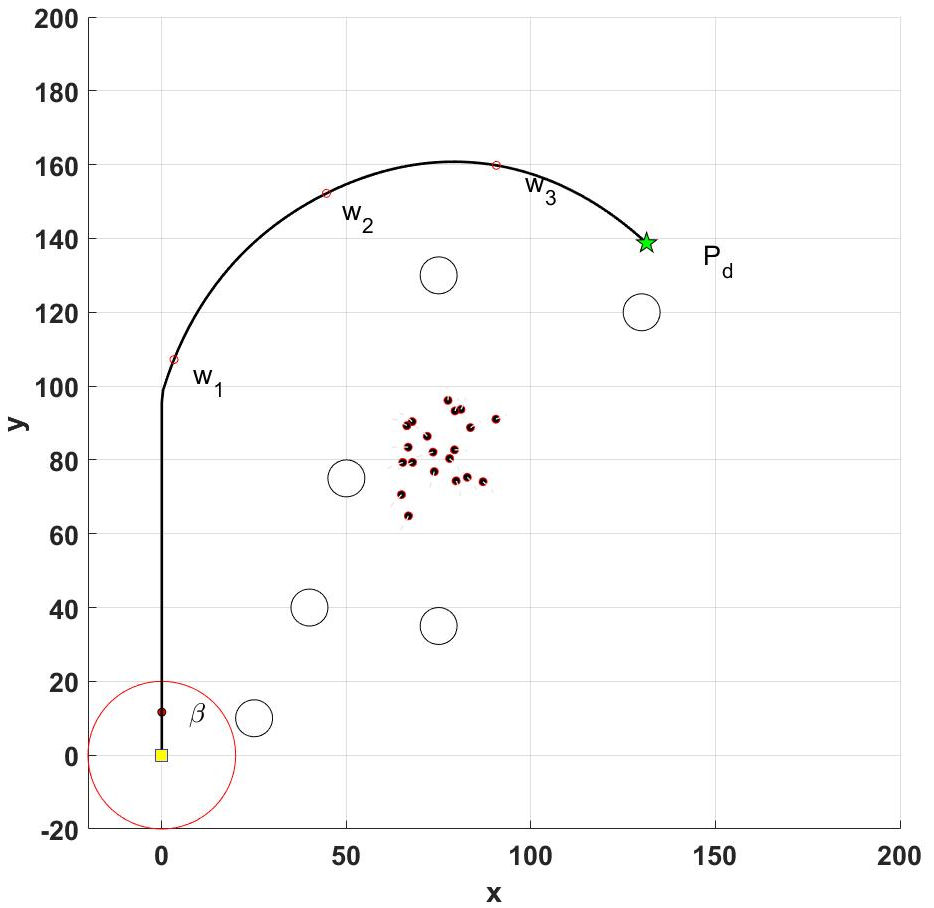}
     \includegraphics[width=0.5\textwidth]{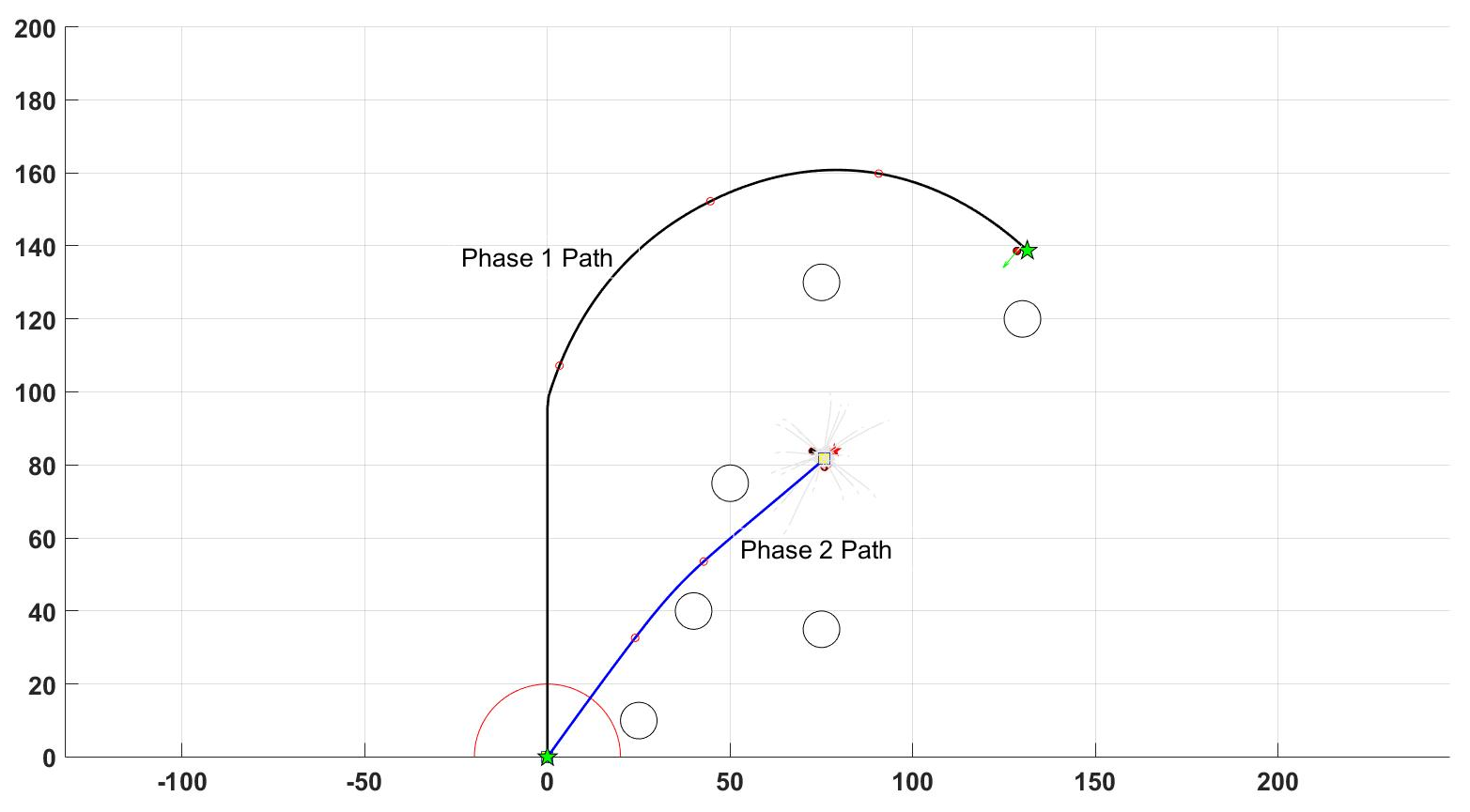}
    \caption{\textbf{Top}: Optimising the path from sheepdog to driving point. Note that every sheep is handled as an obstacle (not shown in this figure) with a radius of 60 units. The  path  is  constrained  to be  obstacle  free and  should  not impact  the  sheep
    \textbf{Bottom}: An optimised path from the collected flock to the goal.}
    \label{fig:stage1}
\end{figure}

The proposed framework begins by determining the driving position $P_{d_j}=GCM+r_{a}$  behind the flock  in the direction to the goal, where $r_a$  is the shepherd influence range. Once  $P_{d_j}$ is determined, the best possible path to move the shepherd from its current position to $P_{d_j}$ is calculated using the proposed DE path planning algorithm, see section \ref{DE}. For this step, each sheep is considered as an obstacle with its radius $R_{pi_i}$ equal to the repulsion from the sheepdog, this allows a safety zone around each sheep, hence no dispersion occurs. The DE algorithm returns the best possible path $A^*_{\beta}$ with $D$ way-points calculated using spline  interpolation as follows:
\begin{equation}\label{eq:optimal_path}
A^*_{\beta}=[w^1_{\beta}, w^2_{\beta}, ..., w^d_{\beta}, ..., w^D_{\beta}]
\end{equation}
To guarantee a smooth path, spline interpolation is used to generate points between these way-points. The shepherd then traverses the way-points generated to reach its driving position (see Figure \ref{fig:stage1}).  

Once the shepherd arrives at the driving point, a safety zone circle $C_F$ is generated around the flock centred at its $GCM$. A second optimisation cycle begins to find the best path from the flock $GCM$ to target position whilst ensuring no  obstacles are within $C_F$.  Algorithm~\ref{alg:framework} summarises the proposed framework and such a path is depicted in Figure \ref{fig:stage1}. 

\subsection{Differential Evolution Based Path Planning Algorithm} \label{DE}

Our approach utilises Differential Evolution (DE) for optimization of the location of the way points. The process starts with a random initial population of size $PS$. Each solution represents a path $A=[w^1, w^2, ..., w^d, ..., w^D]$ with $D$ way-points, each containing its $x$ and $y$ coordinates, allowing for representation as a two-dimensional array. For presentation simplicity, the DE steps discussed below consider only the $x$ values, yet the evolutionary steps are applied to the $y$-coordinates as well. We initialise our possible solutions, $\overrightarrow{x}_z (\forall z=1,2,...,PS)$ within the search space, such that

\begin{equation}
x_{z,j}=\underline{x}_{j}+rand_{j}(0,1)\times(\overline{x}-\underline{x}),\:\forall\,j=1,2,...D\label{eq:initialization}
\end{equation}

\noindent where $rand_{j}(0,1)$ is a uniform random number within $[0,1]$, $\underline{x}$ and $\overline{x}$ are the lower and upper boundaries of the search space, here set to zero and paddock size, respectively).

Each individual is evaluated based on both its objective function and adherence to constraints. At each population evaluation, the number of current fitness evaluations $(cfe)$ is increased. Then, the entire population is evolved by DE operators (mutation and crossover); that is, for every solution $\overrightarrow{x}_{z}$, a new trial individual, $\overrightarrow{u}_{z}$ is generated using:

\begin{equation}
u_{z,j}=\begin{cases}
x_{z,j}+F_{z}.(x_{\phi,j}-x_{z,j}+x_{r_{1},j}-x_{r_{2},j})\\
\hfill{if(rand\leq cr_{z}\,\text{or}\,j=j_{rand})}\\
x_{z,j}\hfill{\text{otherwise}}
\end{cases}\label{eq:DE2}
\end{equation}

\noindent where $F_{z}$ is the amplification factor, $cr_{z}$ is the crossover rate,  $r_{1}\neq r_{2}\neq z$ are random integer numbers, with $\overrightarrow{x}_{r_{1}}$ and $\overrightarrow{x}_{r_{2}}$ randomly selected from $X$, $x_{\phi,j}^{i}$ was selected from the best $10\%$ individuals in $X$ \cite{zhang2009jade}. Note if $u_{z,j}$ violates the search boundary, it is rounded back to the limit, i.e., if $u_{z,j}< \underline{x}$ then $u_{z,j} \leftarrow \underline{x}$, and if  $u_{z,j}> \overline{x}$, then $u_{z,j} \leftarrow \overline{x}$. For a discussion on how we adapt $F_z$ and $Cr_{z}$, please see subsection~\ref{subsec:Adaptation-of-} below.

Once the new solutions are generated, pairwise comparison between every   $\overrightarrow{x}_{z}$ and $\overrightarrow{u}_{z}$ is conducted, with the winner progressing to the next generation; that is  $\overrightarrow{u}_{z}$ survives to the next population if (1) both solutions are feasible, and $f(\overrightarrow{u}_{z}) \leq f(\overrightarrow{x}_{z})$; or (2) $\overrightarrow{u}_{z}$ is feasible, but $\overrightarrow{x}_{z}$ is not; or (3) both solutions are infeasible, and $\psi(\overrightarrow{u}_{z}) \leq \psi(\overrightarrow{x}_{z})$, where the violation of the $z^{th}$ solution is calculated using equation \ref{eq:vio}. The algorithm continues until the stopping criterion is satisfied.

\subsection{Evaluation}
As previously mentioned, the quality of each solution is determined by its fitness value and any constraint violations. In this paper, the objective function considered is the length of a path $(L)$ from the start point to the target location. The following steps are carried out for every solution $\overrightarrow{x_{z}}$.

\begin{enumerate}
    \item set the $x$-coordinate vector $XS\leftarrow[\underline{x},x_{z,1},...x_{z,D},\bar{x}]$;
    \item set the $y$-coordinate vector $YS\leftarrow[\underline{y},y_{z,1},...y_{z,D},\bar{y}]$;
    \item $TS\leftarrow$ split a line into $k=D+2$ points (i.e., equal intervals);
    \item $LS\leftarrow$ generate a vector of $p_{max}=100$ evenly spaced points between $0$ and $1$;
    \item interpolate $XS$ and $YS$ over unevenly-spaced sample points; i.e.,
    generate a vector of interpolated values ($\overrightarrow{XI}$)
    corresponding to the points in $LS$. The values are determined
    by cubic spline interpolation of $TS$ and $XS$. Similarly, a vector
    of interpolated values ($\overrightarrow{YI}$) corresponding to the
    points in $LS$ is generated by the cubic spline interpolation of
    $TS$ and $YS$. Note that the combination of $[\overrightarrow{XI};\overrightarrow{YI}]$
    represents the path from start to target, with each pair of ${XI}_{p}$
    and ${YI}_{p}$, where $p=1,2,...,  p_{max}$  represents a point in this path. 
    \item Subsequently, the length $(L)$ of this path is calculated as follows:
    \begin{equation}
        L_z=\sum_{p=1}^{p_{max}-1}\sqrt{(XI_{p+1}-XI_{p})^{2}+(YI_{p+1}-YI_{p})^{2}}
    \end{equation}
\end{enumerate}

Hence, the objective function is to 
 \begin{equation}
        \text{minimise\,\,\,} L
 \end{equation}

A path is considered feasible if it does not cross any of the obstacles.
Mathematically, the following equation is used to calculate the total constraint 
violation (how much a path overlaps with all obstacles), with a value
of $0$ means no overlapping exist. 
\begin{enumerate}
\item calculate the Euclidean distance between each point in the path and
the centre of each obstacle $(O)$, such that $d_{\ensuremath{p,s}}=\sqrt{(XI_{p}-O_{x,s})^{2}+(YI_{p}-O_{y,s})^{2}}$,
where $O_{x,s}$ and $O_{y,s}$ are the $x$ and $y$ coordinates
of the centre of $O$.
\item the violation of the $z^{th}$ solutions is calculated as 
 \begin{equation} 
 \psi_{z}=\sum_{p=1}^{p_{max}}\sum_{s=1}^{N}\text{maximum}\left(\text{\ensuremath{1-\frac{d_{p,s}}{O_{s,radius}}}},0\right) \label{eq:vio}
 \end{equation}
\noindent where $O_{s,radius}$ is the radius of the $s^{th}$ obstacle. 
 
During the driving behaviour, we intend to move the sheep flock considering it as a circle, so the violation is considered as 
 
 \begin{equation} 
 \psi_{z}=\sum_{p=1}^{p_{max}}\sum_{s=1}^{N}\text{maximum}\left(\text{\ensuremath{1-\frac{d_{p,s}-flock_{radius}}{O_{s,radius}}}},0\right) \label{eq:vio2},
 \end{equation}
  where $flock_{radius}$ is the distance between the global centre of mass and the furthest sheep. 
\end{enumerate}

\subsection{Adaptation of Amplification Factor and Crossover Rate  \label{subsec:Adaptation-of-}} 

Within our implementation of DE, the technique used to adapt $F$ and $Cr$ considering the presence of constraints is based on the work presented in~\cite{elsayed2016enhanced}; that is: 
\begin{enumerate}
\item A memory archive of size $H$ for both parameters ($M_{Cr}$,
$M_{F}$) is set to a value of $0.5$. 
\item Every $\overrightarrow{x}_{z}$ is assigned with its $Cr_{z}$ and $F_{z}$ 
\begin{equation}
Cr_{z}=\text{randni}(M_{Cr,r_{z}},\sigma_{cr})
\end{equation}
\begin{equation}
F_{z}=\text{randci}(M_{F,r_{z}},\sigma_{F})
\end{equation}
where $r_{z}$ is randomly chosen from $\left[1,H\right]$. $\text{randni}(\mu,\sigma)$
and $\text{randci}(\mu,\sigma)$ are generated using 
the normal and Cauchy distributions with mean $\mu$ and variance
$\sigma$, $\sigma_{F}=\sigma_{cr}=0.1$. 

\item After every evolutionary iteration, successful $Cr_{z}$ and $F_{z}$ are recorded in $S_{Cr}$ and $S_{F}$;  the memory archive is updated as follows: 
\begin{equation}
M_{Cr,d}=mean_{WA}\left(S_{Cr}\right)\,if\,S_{Cr}\neq\text{null}
\end{equation}
\begin{equation}
M_{F,d}=mean_{WL}\left(S_{F}\right)\,if\,S_{F}\neq\text{null}
\end{equation}
where $1\le d\le H$ is the position in the memory archive. Note that $d$ is initialised to $1$ and increased by 1 after every addition to the memory archive. If $d>H$, then $d=1$. $\text{mean}_{WA}(S_{Cr})$ and $\text{mean}_{WL}(S_{F})$
are calculated using the following equations:

\begin{equation}
\text{mean}_{WA}(S_{Cr})=\sum_{\gamma=1}^{\left|S_{Cr}\right|}w_{\gamma}.S_{cr,\gamma}
\end{equation}

\begin{equation}
\text{mean}_{WL}(S_{F})=\frac{\sum_{\gamma=1}^{\left|S_{F}\right|}w_{\gamma}.S_{F,\gamma}^{2}}{\sum_{\gamma=1}^{\left|S_{F}\right|}w_{\gamma}.S_{F,\gamma}}
\end{equation}
where

\begin{equation}
w_{\gamma}=\frac{\xi_{\gamma}}{\sum_{\gamma=1}^{\left|S_{Cr}\right|}\xi_{\gamma}}\label{eq:w_zd}
\end{equation}

and $\xi_{\gamma}$ is calculated as follows:

\begin{itemize}
\item First, three scenarios are defined to classify solutions:
\begin{enumerate}
\item  scenario-1 (\textbf{\textit{Infeasible to infeasible}}): $\overrightarrow{x}_{z,t}$ and  $\overrightarrow{x}_{z,t-1}$  are  infeasible, where $t$ is the current generation. 
\item scenario-2 (\textbf{\textit{Infeasible to feasible}}): $\overrightarrow{x}_{z,t-1}$ is infeasible while $\overrightarrow{x}_{z,t}$ is feasible. 
\item scenario-3 (\textbf{\textit{Feasible to feasible}}): $\overrightarrow{x}_{z,t}$ and  $\overrightarrow{x}_{z,t-1}$  are  both feasible. 
\end{enumerate}

\item For each successful solution $\left(\gamma\in1,2,...\left|S_{Cr}\right|\right)$\footnote{$\left|S_{Cr}\right|$ is the number of successful $Cr$ recorded
in $S_{Cr}$, and $\left|S_{Cr}\right|$=$\left|S_{F}\right|$}which satisfies scenario $1$ conditions, the corresponding $I_{\gamma}$ is 
\begin{multline}
\xi_{\gamma}=I_{\gamma}=\mbox{max}\left(0,\frac{\psi_{\gamma,t-1}-\psi_{\gamma,t}}{\psi_{\gamma,t-1}}\right)\\
+\mbox{max}\left(0,\frac{f_{\gamma,t-1}-f_{\gamma,t}}{\left|f_{\gamma,t-1}\right|}\right)\label{eq:I2I- Cr and F}
\end{multline}
\item For those solutions $\left(\gamma\in1,2,...\left|S_{Cr}\right|\right)$
which belong to scenario $2$ or $3$,  $I_{\gamma}$
 
\begin{multline}
\xi_{\gamma}=\mbox{max}\left(0,I_{\gamma}\right)+\frac{\psi_{\gamma,t-1}-\psi_{\gamma,t}}{\psi_{\gamma,t-1}}\\
+\mbox{max}\left(0,\frac{f_{\gamma,t-1}-f_{\gamma,t}}{\left|f_{\gamma,t-1}\right|}\right)\label{eq:I2I- Cr and F-1}
\end{multline}
\end{itemize}
\end{enumerate}

\section{Experimentation and Results}\label{sec:experiments}
\subsection{Str\"{o}mbom vs UNSWDST in Obstacle Free Environment}
In this subsection, we compare Str\"{o}mbom et al\textquoteright against the UNSWDST variant in an obstacle free environment. Using Matlab, we create an environment of size 500x500 units, with different herd sizes comprising 10, 50 and 100 sheep. Across 10 simulation runs for each of the herd sizes, the two algorithms are compared in terms of their ability to effectively (success rate) and efficiently (speed) complete the task as shown in Table~\ref{tab:unswdst}.

The results of this experiment show that both approaches can achieve 100\% success rates with different number of sheep. However, UNSWDST method outperformed the standard Str\"{o}mbom method in terms of number of steps taken to drive the flock to the designated goal area in all three scenarios. Particularly, with small flock sizes UNSWDST can achieve up to 44\% improvement on average. The results highlight also the performance stability of both methods with UNSWDST having lower standard deviation (11.4 to 30\% improvement over Str\"{o}mbom). 

The advantage of UNSWDST algorithm is during collection. This advantage starts to disappear as the flock size increases. The larger flock size comes with a bias, where the total attraction force acting on sheep due to attraction to CoM is high. The flock tends to cluster more, thus, it is less likely to have a sheep requiring collection. This is evident in the case of a flock size of 100 sheep, where Str\"{o}mbom et al.\textquoteright s performance starts to approach that of the UNSWDST algorithm.

\begin{table}[t]
\small
\caption {A comparison of Str\"{o}mbom et al against the UNSWDST algorithm in an obstacle free environment with different herd size.} \label{tab:unswdst}
 \begin{center}
  \begin{tabular}{cccc}
    \hline  \hline
    No. of    & \multirow{2}{*}{Metric} &  UNSWDST & Str\"{o}mbom et al.  \\ 
    sheep   &  & Model  & Model\\
    \hline  \hline
    \multirow{3}{*}{10}   &  best   &  370
   &  512\\
        &  mean   &  410.4
   &  732.5\\
   &  std   &  33.86
   &  112.81\\  \hline
\multirow{3}{*}{50}   &  best   &  373
   &  391
\\
   &  mean   &  376.3
   &  447.1
\\
   &  std   &  3.49
   &  30.61
\\  \hline
\multirow{3}{*}{100}   &  best   &  368
   &  372
\\
   &  mean   &  371.6
   &  378.9
\\
   &  std   &  1.80
   &  6.20
\\  \hline \hline
\end{tabular}
\end{center}
\end{table}

\subsection{Two-Phase Path Planning Framework in Cluttered Environments}

\begin{figure}
    \centering
\includegraphics[width=0.45\textwidth]{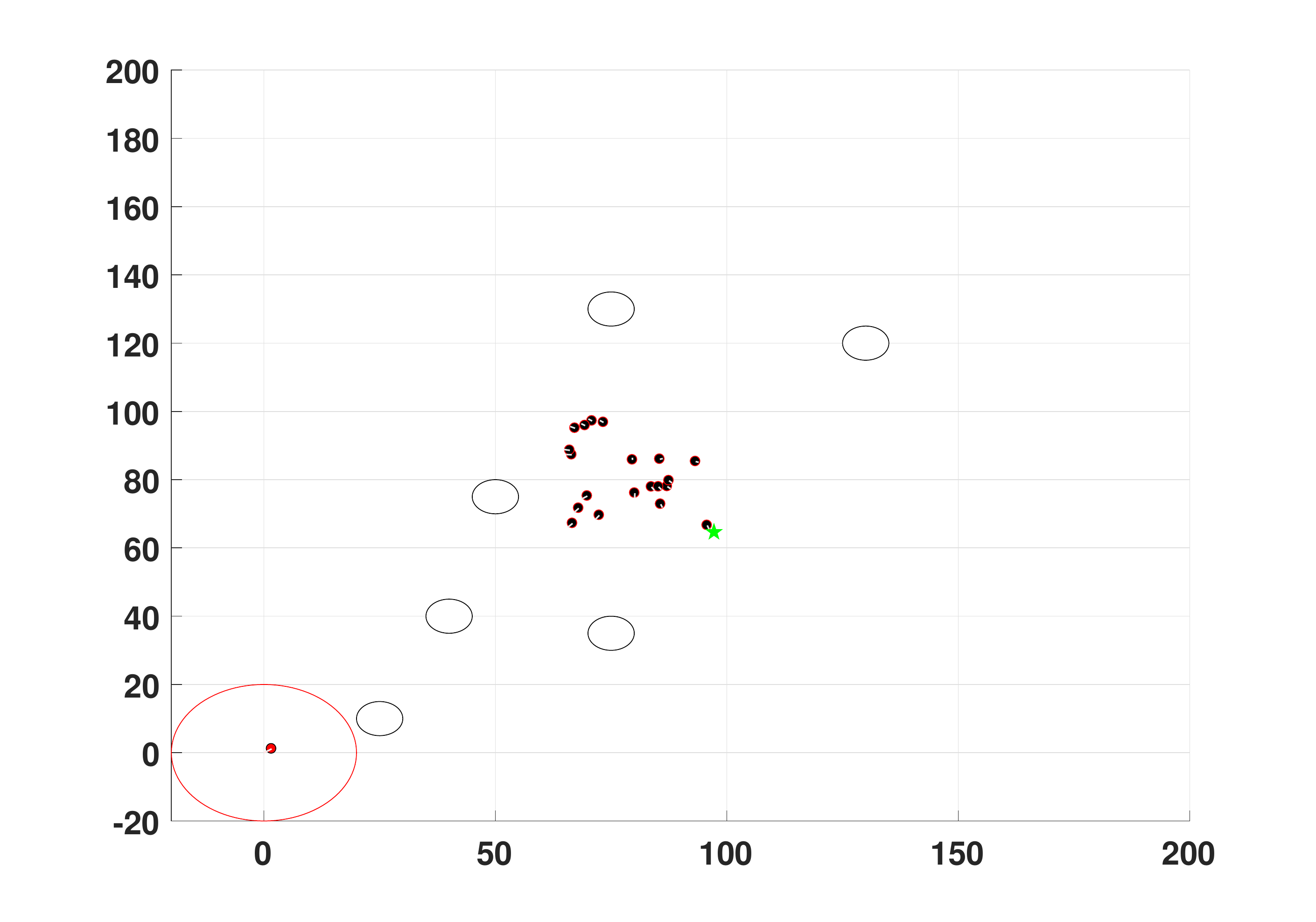}
    
    \includegraphics[width=0.45\textwidth]{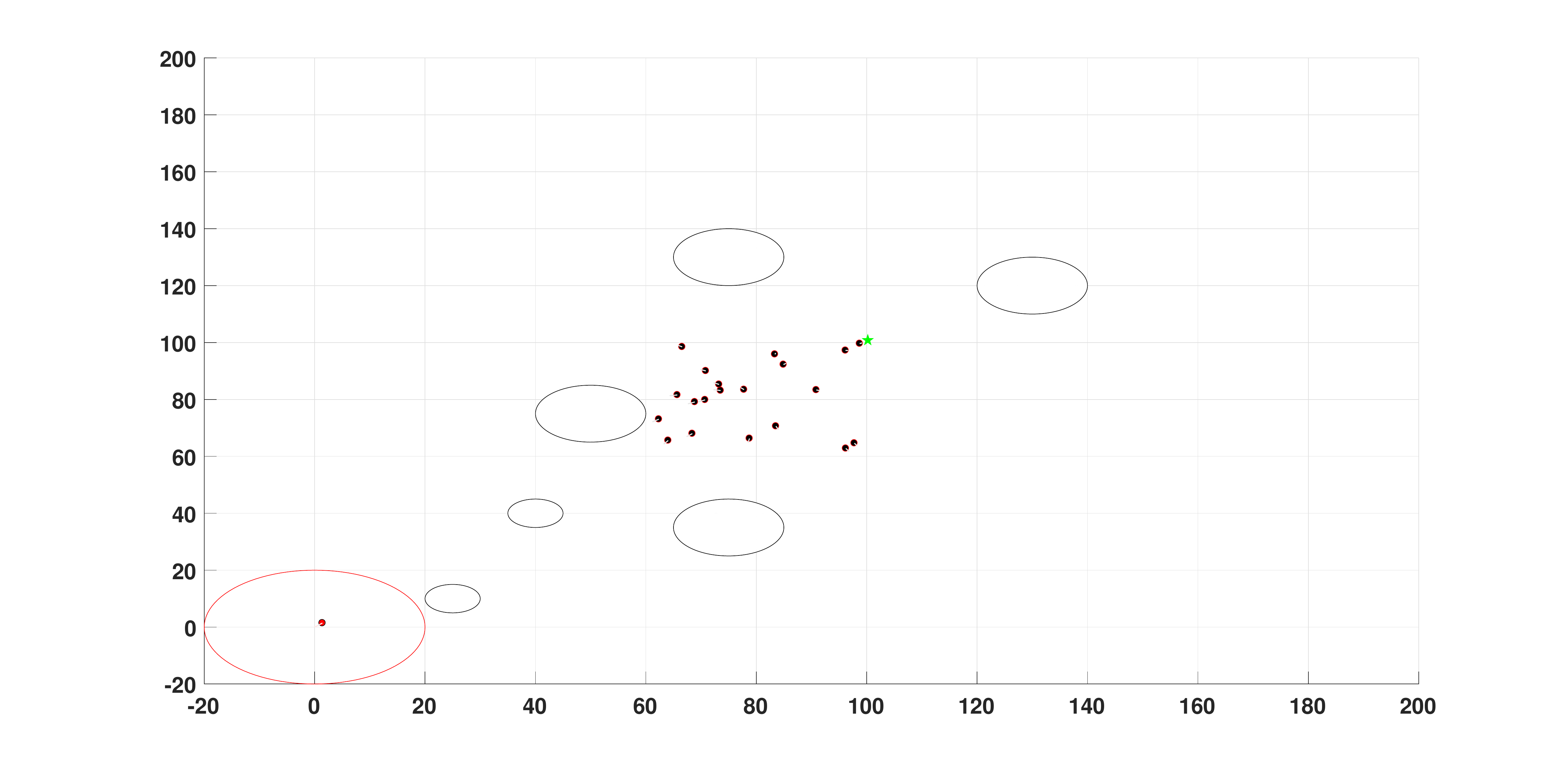}
    
    \includegraphics[width=0.45\textwidth]{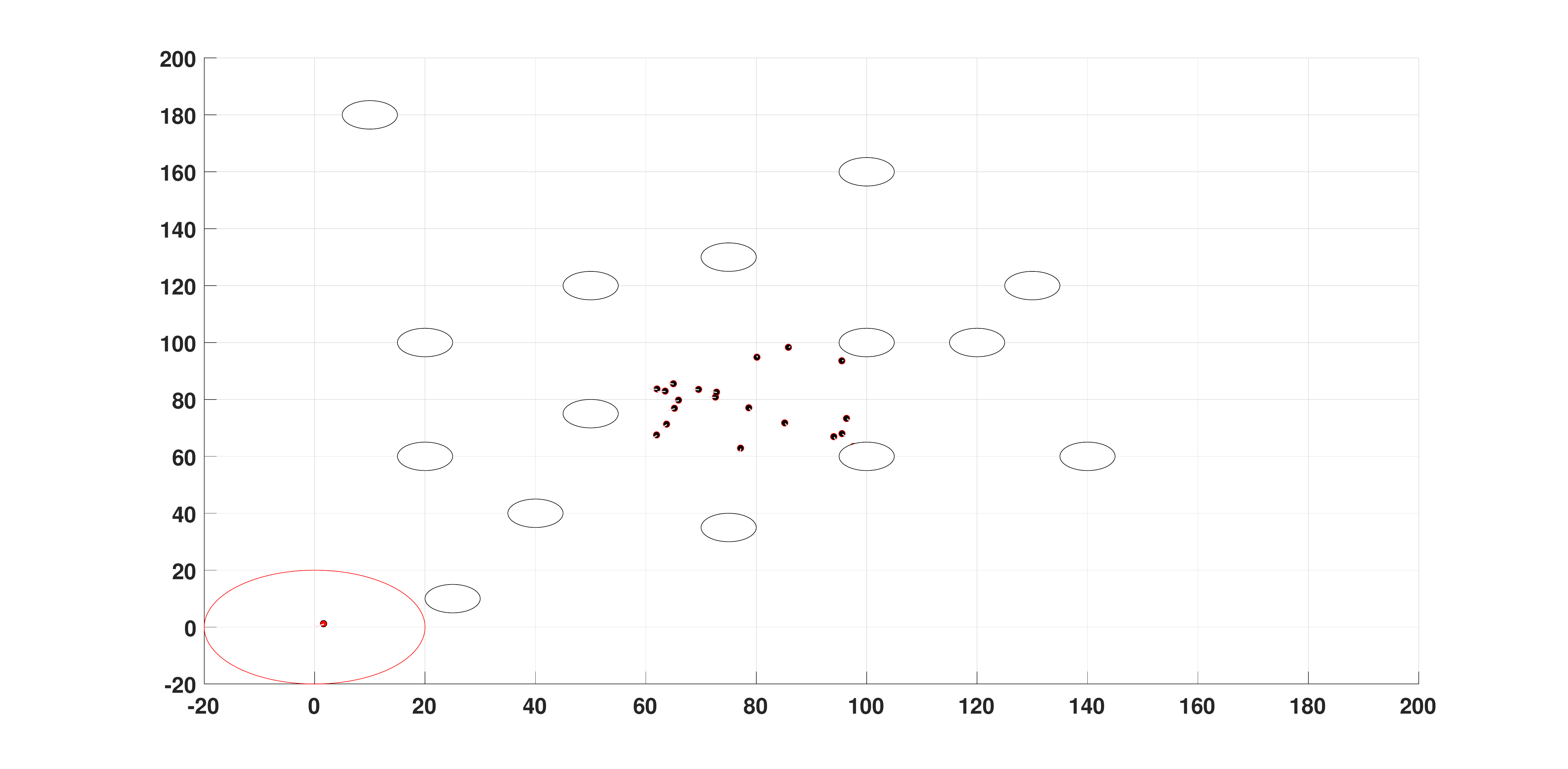}
    \caption{The three scenario configurations with different number of obstacles and obstacle sizes.}
    \label{fig:scenario1}
\end{figure}

Hypothesising that further improvements can be had through the use of path planning, in this subsection, we analyse the performance of the two-phase path planning framework in improving the shepherding task in cluttered environments and compare our results with the newly developed UNSWDST model. 
Our experimental environment, implemented in Matlab, allowed for the comparison to be carried out across multiple scenarios. The environment size is 200x200 units; goal location for shepherding is [0,0] as is the initial shepherd location. The flock is randomly spawned with both the $x$ and $y$ locations are $\geq 60$ and $\leq 100$. Obstacle number is either 6 or 13. Their locations are positioned in the area out the initial flock locations, and their locations are fixed over all runs of the simulation. Obstacle size is set to either small (radius=5) or large  (radius=10).  The flock size has three settings: 20, 40 or 80 sheep. A depiction of the different scenarios is shown in Figure~\ref{fig:scenario1}.

The Differential evolution utilised a population size of 30 individuals with the algorithm running for 150 generations. DE\textquoteright s parameters, $F$ and $CR$ are self-adaptively updated, as per the discussion in section \ref{subsec:Adaptation-of-}. The number of way-points for the path generation was set to $D=3$. Each scenario configuration was run 20 times, with the best, average and standard deviation of the time taken (number of simulation steps) to successfully complete the shepherding task reported in Table~\ref{tab:results}. 

\begin{table}[t]
\small
\caption {Simulation results comparing task completion time} \label{tab:results}
 \begin{center}
  \begin{tabular}{ccccc}
    \hline  \hline
    No. of & No. of    & \multirow{2}{*}{Metric} & UNSWDST  & UNSWDST1\\ 
    obstacles & sheep   &  & Model  & Model\\
    \hline  \hline
    6   &  \multirow{3}{*}{20}   &  best   &  267   &  185\\
   (small) &     &  mean   &  286.95   &  211.65\\
   &     &  std   &  14.31   &  17.29\\  \hline
6   &  \multirow{3}{*}{40}   &  best   &  257   &  191\\
   (small) &     &  mean   &  284.10   &  198.85\\
   &     &  std   &  13.30   &  8.40\\  \hline
6   &  \multirow{3}{*}{80}   &  best   &  256   &  192\\
   (small) &     &  mean   &  339.30   &  199.15\\
   &     &  std   &  152.02   &  5.78\\  \hline
6   &  \multirow{3}{*}{20}   &  best   &  263   &  200\\
   (large) &     &  mean   &  286.40   &  254.60\\
   &     &  std   &  12.16   &  105.40\\  \hline
6   &  \multirow{3}{*}{40}   &  best   &  263   &  196\\
   (large) &     &  mean   &  284.80   &  212.90\\
   &     &  std   &  16.32   &  23.06\\  \hline
6   &  \multirow{3}{*}{80}   &  best   &  259   &  194\\
   (large) &     &  mean   &  308.30   &  254.90\\
   &     &  std   &  62.51   &  104.93\\  \hline
13   &  \multirow{3}{*}{20}   &  best   &  214   &  190\\
   (large) &     &  mean   &  259.55   &  215.35\\
   &     &  std   &  23.44   &  19.44\\  \hline
13   &  \multirow{3}{*}{40}   &  best   &  217   &  191\\
   (large) &     &  mean   &  243.90   &  222.25\\
   &     &  std   &  16.38   &  37.49\\  \hline
13   &  \multirow{3}{*}{60}   &  best   &  223   &  190\\
   (large) &     &  mean   &  247.45   &  218.75\\
   &     &  std   &  20.51   &  18.82\\  \hline
 \hline
\end{tabular}
\end{center}
\end{table}

The reader will note that in all cases investigated, the addition of our two phase DE path planner greatly improves the results over the UNSWDST approach. There are, however, two scenarios where a larger standard deviation was observed. We hypothesise that this variation is caused by the placement of obstacles which challenges the DE to find appropriate obstacle free paths given the low fidelity $D=3$ way-points. Future work will conduct a sensitivity analysis on the number of way-points to allow for more complex path generation.

\section{Conclusion and Future Work}\label{sec:conclusion}

The guidance of a large group of agents (a swarm) using a single point of control couples the dynamics of the control point with that of the group, and constrains the movement of the control point to that feasible for itself and the group as a whole. Within the context of shepherding, we presented a differential evolution algorithm to plan the path for the sheepdog while constraining the path to those appropriate for the sheep it herded. 

Our framework decomposed the problem into path planning to approach the flock (without inducing scattering), and then path planning, via the generation of sub-goal waypoints, to drive the flock whilst minimising encountered obstacles. The efficacy of the approach was evaluated via simulation in environments of varying flock size, obstacle number and obstacle size. 

To evaluate the path planning approach, we modified the Str\"{o}mbom approach to offer a more efficient algorithm for swarm guidance. The modified algorithm without path planning formed the baseline in all scenarios. This modified algorithm was then coupled with the differential evolution based path planning algorithm which provided further gains.

This study opens new interesting questions which require further work. Evaluating our proposed algorithm in different scenarios, such as multiple sheepdogs and dispersed sheep flocks will form some of our future work in this area. Moreover, relaxing the constraint that sheep are handled as obstacles with a large safety zone radius will be evaluated. A comparative study with other evolutionary algorithms would equally be useful.

\section*{Acknowledgement}
The authors would like to acknowledge a US Office of Naval Research - Global (ONR-G) Grant.

\balance


\begin{thebibliography}{10}
\providecommand{\url}[1]{#1}
\csname url@samestyle\endcsname
\providecommand{\newblock}{\relax}
\providecommand{\bibinfo}[2]{#2}
\providecommand{\BIBentrySTDinterwordspacing}{\spaceskip=0pt\relax}
\providecommand{\BIBentryALTinterwordstretchfactor}{4}
\providecommand{\BIBentryALTinterwordspacing}{\spaceskip=\fontdimen2\font plus
\BIBentryALTinterwordstretchfactor\fontdimen3\font minus
  \fontdimen4\font\relax}
\providecommand{\BIBforeignlanguage}[2]{{%
\expandafter\ifx\csname l@#1\endcsname\relax
\typeout{** WARNING: IEEEtran.bst: No hyphenation pattern has been}%
\typeout{** loaded for the language `#1'. Using the pattern for}%
\typeout{** the default language instead.}%
\else
\language=\csname l@#1\endcsname
\fi
#2}}
\providecommand{\BIBdecl}{\relax}
\BIBdecl

\bibitem{long2020comprehensive}
N.~K. Long, K.~Sammut, D.~Sgarioto, M.~Garratt, and H.~A. Abbass, ``A
  comprehensive review of shepherding as a bio-inspired swarm-robotics guidance
  approach,'' \emph{IEEE Transactions on Emerging Topics in Computational
  Intelligence}, pp. 523--537, 2020.

\bibitem{lien2009interactive}
J.-M. Lien and E.~Pratt, ``Interactive planning for shepherd motion.'' in
  \emph{AAAI Spring Symposium: Agents that Learn from Human Teachers}, 2009,
  pp. 95--102.

\bibitem{fingas2012basics}
M.~Fingas, \emph{The basics of oil spill cleanup}.\hskip 1em plus 0.5em minus
  0.4em\relax CRC press, 2012.

\bibitem{shell2004directional}
D.~A. Shell and M.~J. Mataric, ``Directional audio beacon deployment: an
  assistive multi-robot application,'' in \emph{IEEE International Conference
  on Robotics and Automation, 2004. Proceedings. ICRA'04. 2004}, vol.~3.\hskip
  1em plus 0.5em minus 0.4em\relax IEEE, 2004, pp. 2588--2594.

\bibitem{chaimowicz2007aerial}
L.~Chaimowicz and V.~Kumar, ``Aerial shepherds: Coordination among uavs and
  swarms of robots,'' in \emph{Distributed Autonomous Robotic Systems 6}.\hskip
  1em plus 0.5em minus 0.4em\relax Springer, 2007, pp. 243--252.

\bibitem{patle2019review}
B.~Patle, G.~B. L], A.~Pandey, D.~Parhi, and A.~Jagadeesh, ``A review: On path
  planning strategies for navigation of mobile robot,'' \emph{Defence
  Technology}, vol.~15, no.~4, pp. 582 -- 606, 2019.

\bibitem{reynolds1987flocks}
C.~W. Reynolds, ``Flocks, herds and schools: A distributed behavioral model,''
  in \emph{Proceedings of the 14th annual conference on Computer graphics and
  interactive techniques}, 1987, pp. 25--34.

\bibitem{vaughan1997introducing}
R.~Vaughan, J.~Henderson, and N.~Sumpter, ``Introducing the robot sheepdog
  project,'' in \emph{Proceedings of the International Workshop on Robotics and
  Automated Machinery for BioProductions}, 1997.

\bibitem{vaughan2000experiments}
R.~Vaughan, N.~Sumpter, J.~Henderson, A.~Frost, and S.~Cameron, ``Experiments
  in automatic flock control,'' \emph{Robotics and autonomous systems},
  vol.~31, no. 1-2, pp. 109--117, 2000.

\bibitem{el2020preliminary}
H.~El-Fiqi, B.~Campbell, S.~Elsayed, A.~Perry, H.~K. Singh, R.~Hunjet, and
  H.~Abbass, ``A preliminary study towards an improved shepherding model,'' in
  \emph{Proceedings of the 2020 Genetic and Evolutionary Computation Conference
  Companion}, 2020, pp. 75--76.

\bibitem{strombom2014solving}
D.~Str{\"o}mbom, R.~P. Mann, A.~M. Wilson, S.~Hailes, A.~J. Morton, and D.~JT,
  ``Solving the shepherding problem: heuristics for herding,'' \emph{Journal of
  The Royal Society Interface}, 2014.

\bibitem{vseda2007roadmap}
M.~{\v{S}}eda, ``Roadmap methods vs. cell decomposition in robot motion
  planning,'' in \emph{Proceedings of the 6th WSEAS international conference on
  signal processing, robotics and automation}.\hskip 1em plus 0.5em minus
  0.4em\relax World Scientific and Engineering Academy and Society (WSEAS),
  2007, pp. 127--132.

\bibitem{glavavski2009robot}
D.~Glava{\v{s}}ki, M.~Volf, and M.~Bonkovic, ``Robot motion planning using
  exact cell decomposition and potential field methods,'' in \emph{Proceedings
  of the 9th WSEAS international conference on Simulation, modelling and
  optimization}.\hskip 1em plus 0.5em minus 0.4em\relax World Scientific and
  Engineering Academy and Society (WSEAS), 2009, pp. 126--131.

\bibitem{bhattacharya2008roadmap}
P.~Bhattacharya and M.~L. Gavrilova, ``Roadmap-based path planning-using the
  voronoi diagram for a clearance-based shortest path,'' \emph{IEEE Robotics \&
  Automation Magazine}, vol.~15, no.~2, pp. 58--66, 2008.

\bibitem{wein2007visibility}
R.~Wein, J.~P. Van~den Berg, and D.~Halperin, ``The visibility--voronoi complex
  and its applications,'' \emph{Computational Geometry}, vol.~36, no.~1, pp.
  66--87, 2007.

\bibitem{khatib1986real}
O.~Khatib, ``Real-time obstacle avoidance for manipulators and mobile robots,''
  in \emph{Autonomous robot vehicles}.\hskip 1em plus 0.5em minus 0.4em\relax
  Springer, 1986, pp. 396--404.

\bibitem{huang2008velocity}
L.~Huang, ``Velocity planning for a mobile robot to track a moving target - a
  potential field approach,'' \emph{Robotics and Autonomous Systems}, vol.~57,
  no.~1, pp. 55 -- 63, 2009.

\bibitem{hoy2015algorithms}
M.~Hoy, A.~S. Matveev, and A.~V. Savkin, ``Algorithms for collision-free
  navigation of mobile robots in complex cluttered environments: a survey,''
  \emph{Robotica}, vol.~33, no.~3, pp. 463--497, 2015.

\bibitem{shing1993genetic}
M.-T. Shing and G.~B. Parker, ``Genetic algorithms for the development of
  real-time multi-heuristic search strategies.'' in \emph{ICGA}.\hskip 1em plus
  0.5em minus 0.4em\relax Citeseer, 1993, pp. 565--572.

\bibitem{shi2010dynamic}
P.~Shi and Y.~Cui, ``Dynamic path planning for mobile robot based on genetic
  algorithm in unknown environment,'' in \emph{2010 Chinese control and
  decision conference}.\hskip 1em plus 0.5em minus 0.4em\relax IEEE, 2010, pp.
  4325--4329.

\bibitem{kang2011genetic}
X.~Kang, Y.~Yue, D.~Li, and C.~Maple, ``Genetic algorithm based solution to
  dead-end problems in robot navigation,'' \emph{International Journal of
  Computer Applications in Technology}, vol.~41, no. 3-4, pp. 177--184, 2011.

\bibitem{castellano1997automatic}
G.~Castellano, G.~Attolico, and A.~Distante, ``Automatic generation of fuzzy
  rules for reactive robot controllers,'' \emph{Robotics and Autonomous
  Systems}, vol.~22, no.~2, pp. 133--149, 1997.

\bibitem{na2003hybrid}
Y.-K. Na and S.-Y. Oh, ``Hybrid control for autonomous mobile robot navigation
  using neural network based behavior modules and environment classification,''
  \emph{Autonomous Robots}, vol.~15, no.~2, pp. 193--206, 2003.

\bibitem{tang2014mobile}
X.-l. Tang, L.-m. Li, and B.-j. Jiang, ``Mobile robot slam method based on
  multi-agent particle swarm optimized particle filter,'' \emph{The Journal of
  China Universities of Posts and Telecommunications}, vol.~21, no.~6, pp.
  78--86, 2014.

\bibitem{liu2019comprehensive}
J.~Liu, S.~Anavatti, and M.~G.~H. Abbass, ``Comprehensive learning particle
  swarm optimisation with limited local search for uav path planning,'' in
  \emph{2019 IEEE Symposium Series on Computational Intelligence (SSCI)}.\hskip
  1em plus 0.5em minus 0.4em\relax IEEE, 2019, pp. 2287--2294.

\bibitem{zhang2013robot}
Y.~Zhang, D.-W. Gong, and J.-H. Zhang, ``Robot path planning in uncertain
  environment using multi-objective particle swarm optimization,''
  \emph{Neurocomputing}, vol. 103, pp. 172--185, 2013.

\bibitem{bayazit2002roadmap}
O.~B. Bayazit, J.-M. Lien, and N.~M. Amato, ``Roadmap-based flocking for
  complex environments,'' in \emph{10th Pacific Conference on Computer Graphics
  and Applications, 2002. Proceedings.}\hskip 1em plus 0.5em minus 0.4em\relax
  IEEE, 2002, pp. 104--113.

\bibitem{lien2004shepherding}
J.-M. Lien, O.~B. Bayazit, R.~T. Sowell, S.~Rodriguez, and N.~M. Amato,
  ``Shepherding behaviors,'' in \emph{IEEE International Conference on Robotics
  and Automation, 2004. Proceedings. ICRA'04. 2004}, vol.~4.\hskip 1em plus
  0.5em minus 0.4em\relax IEEE, 2004, pp. 4159--4164.

\bibitem{vo2009behavior}
C.~Vo, J.~F. Harrison, and J.-M. Lien, ``Behavior-based motion planning for
  group control,'' in \emph{2009 IEEE/RSJ International Conference on
  Intelligent Robots and Systems}.\hskip 1em plus 0.5em minus 0.4em\relax IEEE,
  2009, pp. 3768--3773.

\bibitem{longshepherding}
N.~Long, M.~Garratt, D.~Sgarioto, K.~Sammut, and H.~Abbass, ``Shepherding
  autonomous goal-focused swarms in unknown environments using space-filling
  paths.''

\bibitem{singh2019modulation}
H.~Singh, B.~Campbell, S.~Elsayed, A.~Perry, R.~Hunjet, and H.~Abbass,
  ``Modulation of force vectors for effective shepherding of a swarm: A
  bi-objective approach,'' in \emph{2019 IEEE Congress on Evolutionary
  Computation (CEC)}.\hskip 1em plus 0.5em minus 0.4em\relax IEEE, 2019, pp.
  2941--2948.

\bibitem{zhang2009jade}
J.~Zhang and A.~C. Sanderson, ``Jade: adaptive differential evolution with
  optional external archive,'' \emph{IEEE Transactions on Evolutionary
  Computation}, vol.~13, no.~5, pp. 945--958, 2009.

\bibitem{elsayed2016enhanced}
S.~Elsayed, R.~Sarker, and C.~C. Coello, ``Enhanced multi-operator differential
  evolution for constrained optimization,'' in \emph{2016 IEEE Congress on
  Evolutionary Computation (CEC)}.\hskip 1em plus 0.5em minus 0.4em\relax IEEE,
  2016, pp. 4191--4198.

\end{thebibliography}
\end{document}